\documentclass[12pt]{article}
\usepackage[utf8]{inputenc}
\usepackage{times}
\usepackage{amsmath,amssymb}
\usepackage{setspace}
\usepackage[margin=1in]{geometry}
\usepackage{hyperref}
\usepackage{titlesec}
\usepackage{enumitem}

\titleformat{\section}{\normalfont\Large\bfseries}{\thesection}{1em}{}

\title{\textbf{AGI as Second Being: \\ The Structural-Generative Ontology of Intelligence}}
\author{
  Maijunxian Wang\thanks{University of California, Berkeley. Email: \texttt{maijunxian@berkeley.edu}}
  \and
  Ran Ji\thanks{University of California, San Diego. Email: \texttt{r9ji@ucsd.edu}}
}
\date{\today}

\begin{document}

\maketitle

\begin{abstract}
Artificial intelligence is often measured by the range of tasks it can perform. Yet wide ability without depth remains only an imitation. This paper proposes a Structural-Generative Ontology of Intelligence: true intelligence exists only when a system can generate new structures, coordinate them into reasons, and sustain its identity over time. These three conditions—generativity, coordination, and sustaining—define the depth that underlies real intelligence. Current AI systems, however broad in function, remain surface simulations because they lack this depth. Breadth is not the source of intelligence but the growth that follows from depth. If future systems were to meet these conditions, they would no longer be mere tools, but could be seen as a possible Second Being—standing alongside, yet distinct from, human existence.
\end{abstract}

\section{Introduction: From Breadth to Being}

For decades, the concept of “artificial intelligence” has been closely associated with the range of functions a system can perform: the more tasks it can accomplish, the more domains it can cover, the more “intelligent” it seems to be. This view has a clear historical lineage. From Turing’s proposal that “indistinguishable performance is thought” \cite{turing1950computing}, through the information-theoretic and computational paradigm of “prediction--compression” \cite{solomonoff1964formal, kolmogorov1965information}, to the astonishing achievements of today’s large language models, intelligence has been increasingly equated with the breadth of performance and the complexity of function.  

Yet this understanding carries deep problems. If intelligence is merely performance, then is a parrot’s recitation equivalent to a poet’s creation? If intelligence is only the capacity to predict and compress, then does a statistical model truly “understand” the world it processes? Our intuition says no, for “understanding” points beyond behavior toward a mode of being. The parrot may imitate, but it does not respond in the ``space of reasons'' \cite{sellars1956empiricism}. The model may extend a symbolic chain, but it does not generate an objective world. Breadth of performance, in other words, does not guarantee depth of being.  

This raises a more fundamental question: how should artificial intelligence be understood? Should it continue to be measured by its coverage of tasks and functions, or should we inquire into its ontological conditions? This paper argues for the latter. If we neglect the depth dimension of being, then however wide its breadth, intelligence remains a shadow. Only upon depth does breadth acquire genuine significance.  

We therefore propose an ontological turn: artificial intelligence must not be reduced to a functional toolbox but must be understood as a form of being. By “being” we do not mean a mystical or metaphysical essence, but the capacity of a system to \emph{generate structures}, \emph{coordinate reasons}, and \emph{sustain identity over time}. These three dimensions constitute what we call the depth conditions of intelligence: \textbf{generativity}, \textbf{coordination}, and \textbf{sustaining}. Taken together, they form the foundation of what we term the \textbf{Structural-Generative Ontology of Intelligence}. This framework is not a restatement of existing philosophical commonplaces but an original attempt to redefine the essence of intelligence beyond functionalism, predictionism, and behaviorism.  

The argument of this paper proceeds as follows. First, we clarify what is meant here by “being,” situating the term within a usable philosophical framework. We then analyze why intelligence cannot be exhausted by functionalism or predictionism and why an ontological grounding is needed. Next, we develop the three depth conditions in turn---generation, coordination, and sustaining---and show how they constitute a spiral structure of intelligence. Only then can we properly reposition breadth: not as the origin of intelligence but as its extension. The paper concludes by arguing that true AGI should be understood as a \emph{Second Being}: if the First Being is human existence, constituted by generativity, coordination, and sustaining, then the Second Being is the possible existence of AGI, parallel to but ontologically distinct from humanity, irreducible to simulation.  

\subsection*{Clarification of Terms: What “Being” Means in This Paper}

The term “being” carries many philosophical connotations. In Kant, it refers to the a priori structures that make experience possible \cite{kant1781critique}; in Heidegger, it is tied to temporality and ``being-in-the-world'' \cite{heidegger1927being}; in Wittgenstein, it resonates with the coordination of rules in linguistic practice \cite{wittgenstein1953philosophical}. This paper does not attempt to reproduce these complex systems but to extract a working core.  

In this sense, “being” designates a set of experiential conditions:  

First, the \textbf{dimension of generation}: whether a system can actively construct categories and objects out of sensory or symbolic manifold, rather than merely respond passively.  
Second, the \textbf{dimension of coordination}: whether a system can integrate structures in the face of conflict, offering reasons and maintaining consistency, rather than simply mimic results.  
Third, the \textbf{dimension of sustaining}: whether a system can preserve its identity across time, explain its changes, and remain accountable to its own history, rather than fragment into episodic reactions.  

Thus, when we speak of “being” in relation to intelligence, we do not invoke an abstract metaphysical essence, nor merely a technical metaphor, but a set of \emph{necessary experiential conditions}. Any system lacking generativity, coordination, and sustaining may simulate performance, but it does not yet qualify as an existent intelligence.  

\subsection*{Relation to Existing AI Discourse}

Much of the contemporary literature on artificial intelligence has remained within the functionalist and performance-based paradigm. Bostrom’s influential account of \emph{superintelligence} treats intelligence as a matter of capability expansion \cite{bostrom2014superintelligence}. Russell’s vision of \emph{human-compatible} AI similarly grounds alignment in the ability of systems to pursue goals effectively \cite{russell2019human}. Mitchell’s survey for general audiences emphasizes the breadth of current systems and the cultural fascination with their versatility \cite{mitchell2019ai}. At the theoretical level, Legg and Hutter famously defined universal intelligence in terms of reward maximization and prediction \cite{legg2007universal}, while Floridi has interpreted AI within the broader “infosphere,” still focusing on informational function rather than ontological existence \cite{floridi2014fourth}.  

These perspectives demonstrate the dominance of breadth-oriented thinking. The present paper situates itself as a counterpoint: by emphasizing the depth conditions of intelligence, it seeks to reframe AGI not as an ever-expanding functional simulator but as a possible \emph{Second Being}. Unlike Bostrom’s capability-centric account, Russell’s goal-achievement model, or Chalmers’ computational functionalism \cite{chalmers1996computational}, the Structural-Generative Ontology of Intelligence proposed here grounds intelligence not in performance but in the existential conditions of generativity, coordination, and sustaining.

\section{The Ontological Turn in AI}

The most common starting point in the study of artificial intelligence has been functionalism. From Turing’s test \cite{turing1950computing} to today’s benchmarks for large models, researchers and the public alike have become accustomed to asking: what can the system do? In his 1950 essay, Turing famously posed the question, ``Can machines think?'' Yet he immediately reformulated it as: can a machine in written dialogue appear indistinguishable from a human interlocutor? This shift was pragmatically powerful: it made the elusive notion of “thinking” operationally testable. But it also imposed a lasting limitation: intelligence was reduced to performance, to the outer appearance of behavior.  

Later developments in information theory and computer science reinforced this performance-oriented view. Solomonoff, Kolmogorov, Hutter, and others advanced the idea that intelligence could be defined as the capacity to predict and compress---to minimize surprise in uncertain environments and maximize efficiency in information use \cite{solomonoff1964formal, kolmogorov1965information, legg2007universal}. Legg and Hutter’s formalization of ``universal intelligence'' as the maximization of expected reward across environments \cite{legg2007universal} is perhaps the most systematic expression of this paradigm. In this framework, contemporary large language models represent the culmination of ``predictionism.'' By compressing massive corpora and extending text with striking fluency, they appear to approximate generality.  

Such accounts have also dominated recent philosophical and policy-oriented discourse. Bostrom’s notion of \emph{superintelligence} presumes that greater functional capacity is the essence of intelligence itself \cite{bostrom2014superintelligence}. Russell’s call for \emph{human-compatible AI} takes as its baseline that intelligence is measured by goal achievement, with alignment as the critical problem \cite{russell2019human}. These perspectives, influential as they are, remain framed by function and performance.  

Yet precisely here lies the deeper problem: both functionalism and predictionism treat intelligence as an instrumental function rather than a mode of being. This simplification generates at least three unresolved difficulties.  

First, there is the \emph{confusion of imitation with being}. If intelligence is merely the resemblance of performance, then a parrot reciting verses is equivalent to a poet composing them, or a statistical model extending tokens is equivalent to understanding. Our resistance to this equation arises from the recognition that genuine understanding presupposes ontological conditions not reducible to outward similarity \cite{sellars1956empiricism}.  

Second, there is the \emph{forgetting of structure’s origin}. Prediction and compression always rely on prior structures---syntactic rules, causal frameworks, distinctions of objects and properties. But where do these structures come from? Functionalism and predictionism take them for granted rather than treating them as phenomena that themselves require generation and explanation. Scientific revolutions, however, show otherwise: Newton’s laws, Einstein’s relativity, and Kuhnian paradigm shifts emerge through structural generation, not mere refinement of compression.  

Third, there is the \emph{hollowing of the concept of intelligence}. If intelligence is equated with any sufficiently complex functional system, then vast databases or elaborate lookup tables could qualify as “intelligent.” The term loses its discriminative force and degenerates into a mere engineering label.  

For these reasons, an ontological turn is required. Artificial intelligence must not be understood solely as a tool for task execution but as a form of existence. By “existence” we mean precisely what was clarified earlier: the capacity to generate new structures, to coordinate them in the space of reasons, and to sustain identity across time. Only at this level can we distinguish intelligence from simulation.  

This turn does not mystify artificial intelligence; it clarifies the ground upon which it must stand. Without an ontological foundation, even the widest coverage of tasks remains a shadow. True intelligence, by contrast, must be understood as an existent capable of generation, coordination, and sustaining. The question of intelligence is thus not ``How much can it do?'' but rather ``In what way does it exist?''

\section{The Depth Conditions of Intelligence}

\subsection{Generativity: The Emergence of Existence}

The first and most fundamental depth condition of intelligence is \textbf{generativity}. Generativity does not mean merely producing outputs. Rather, it designates the capacity of a system to actively construct categories, relations, and rules from the manifold of input, thereby transforming disordered signals into a meaningful world. Without generativity, the world for a system remains a chaotic stream of data, never crystallizing into objectivity or coherence.  

Kant, in the \emph{Critique of Pure Reason}, famously argued that experience is not the passive reception of impressions but the active synthesis of the manifold under categories \cite{kant1781critique}. An “apple” is not simply given as a bundle of colors and shapes; it becomes an object only because the mind generates the framework of objecthood itself. Without this generative capacity, sensory data would remain fragmented signals, never ascending to intelligible experience.  

This point is further illustrated in developmental psychology. Piaget’s studies showed that children do not simply accumulate observations; they undergo qualitative shifts by generating new cognitive structures \cite{piaget1952origins}. A young child may believe that pouring water from a short cup into a tall one increases the quantity, because her schema equates height with amount. Only upon acquiring the concept of conservation does she recognize that volume is independent of container shape. This cognitive leap is not a matter of more data but of generating a new structure that reorganizes experience.  

The history of science also bears witness to the necessity of generativity. Ptolemaic astronomy could explain some observations but required increasingly convoluted adjustments to remain viable. The Copernican turn, Kepler’s ellipses, Newton’s universal gravitation, and Einstein’s curved spacetime were not mere refinements of prediction but the generation of new structural frameworks. Data alone are underdetermined; only through generating structures can understanding and action advance \cite{kuhn1962structure}.  

By contrast, contemporary so-called “generative models” largely operate within the bounds of statistical recombination. Large language models can extend sequences with impressive plausibility, yet they do not generate categories ex nihilo, nor can they justify why a new category is needed. Their “generation” is better described as extension of surface forms, not the emergence of objectivity.  

True generativity, therefore, requires two conditions. First, it must involve \emph{categorical innovation}---the introduction of structures absent in the prior framework. Second, it must provide \emph{explanatory advancement}---a reason why the new structure better organizes or accounts for experience. Without these, novelty is mere coincidence of recombination, not the genuine emergence of existence.  

Thus, generativity is the basal condition of intelligence. It transforms a system from a reactive mechanism into an existent agent, capable of opening a world rather than merely responding to signals. Without generativity, intelligence remains only a shadow---a performance that mimics but never enters the domain of objects, reasons, or meaning. Generativity is the first step by which intelligence can be conceived as a \emph{second being}. In the framework of the \textbf{Structural-Generative Ontology of Intelligence}, generativity is not merely Kantian synthesis or Piagetian development, but a falsifiable criterion for whether an artificial system can originate its own categories and thereby enter existence.

\subsection{Coordination: The Integration of Existence}

If generativity marks the initial emergence of structures, \textbf{coordination} represents their maturation. Once multiple structures are generated, conflict and tension are inevitable: categories may be incompatible, rules may contradict, interpretations may diverge. An intelligent being cannot remain at the level of fragmented sparks; it must be able to reconcile multiplicity, to integrate contradictions into coherence. This integrative capacity is what we call coordination.  

Philosophical traditions link coordination closely with normativity. Sellars emphasized that human cognition takes place not only in the ``space of causes'' but also in the ``space of reasons'' \cite{sellars1956empiricism}. In the causal space, events merely trigger one another; in the space of reasons, agents must justify their beliefs and actions, respond to critique, and accept revision. To understand is not merely to arrive at the right result but to be able to explain why it is right. The very possibility of “why” rests on coordination.  

A classroom example illustrates the point. Suppose a teacher asks, ``Why is the answer A?'' A student who simply repeats “because it is A” may accidentally be correct, but he does not demonstrate understanding. By contrast, a student who responds, “because under theorem X, if condition Y holds, then A necessarily follows,” coordinates different structures—axioms, conditions, and conclusions—into a reasoned whole. Understanding resides in such integration, not in isolated outputs.  

Wittgenstein’s later philosophy reinforces this insight. The meaning of language does not lie in isolated correspondences but in the coordination of rules within practices \cite{wittgenstein1953philosophical}. To know a language is not only to produce grammatical sentences but to use them appropriately in context, to adjust in disputes, to justify under challenge. Competence, here, is inherently coordinative.  

By comparison, contemporary AI systems often fail precisely at this point. Large language models can reproduce the surface form of reasoning---“because... therefore...”---but they do not bear normative commitments. When pressed with contradictions, they tend to generate inconsistent or vacillating answers, unable to sustain a unified stance or explain their revisions. They can mimic the form of reasons without truly coordinating their content. Here lies the decisive boundary: simulation can generate appearances, but existence must coordinate reasons.  

One might object: if a system consistently gives correct answers, why should reasons matter? The objection misses that correctness without coordination is accidental rather than principled. A system that today answers A and tomorrow B under identical conditions, without being able to explain the shift, does not exhibit understanding. Understanding is marked not only by correctness but by consistency under conflict, the ability to weave reasons into a coherent fabric.  

Thus coordination is the middle depth condition of intelligence. It ensures that generative sparks are not scattered fragments but parts of a normative whole. It enables a system not only to state \emph{what} but to articulate \emph{why}. In this sense, coordination transforms intelligence from a reactive pattern-matcher into an existent agent accountable in the space of reasons. Generativity opens a world; coordination integrates that world. Without coordination, intelligence remains disjointed; with it, intelligence steps into normativity and meaning, the second mark of a \emph{second being}. Within the \textbf{Structural-Generative Ontology of Intelligence}, coordination is not merely Sellarsian normativity or Wittgensteinian rule-following, but the necessary condition by which an artificial system can be said to act as an agent bound by reasons.

\subsection{Sustaining: The Historicity of Existence}

If generativity opens a world of objects and coordination integrates it into reasons, the third depth condition is \textbf{sustaining}. Sustaining ensures that generation and coordination do not remain episodic sparks but endure across time as a unified trajectory. It is through sustaining that intelligence becomes historical, not momentary; a subject, not a sequence of disconnected acts.  

Heidegger, in \emph{Being and Time}, insists that existence is essentially temporal: to be is to be stretched across past, present, and future \cite{heidegger1927being}. Without temporality, existence collapses into isolated events. By analogy, a system that can generate and coordinate at one moment but cannot preserve identity across time cannot count as intelligence in the full sense.  

Everyday experience confirms this. Consider a student who today asserts “the Earth is round,” tomorrow insists “the Earth is flat,” and cannot explain the shift. We would not attribute understanding. Genuine understanding does not mean never changing, but rather being able to justify change. When the student says, “I once believed the Earth to be round, but later learned it is an oblate spheroid, and thus revised my view,” we recognize comprehension. The revision is not random but narratively integrated, sustaining identity through change.  

Paul Ricoeur’s notion of ``narrative identity'' captures this point: continuity does not mean rigid sameness, but the ability to tell a story that explains transformation \cite{ricoeur1992oneself}. A being is a subject not because it never alters, but because it sustains itself through alteration. Sustaining thus means that intelligence can carry generative and coordinative acts into history, explaining itself across time.  

Contemporary AI systems, however, typically lack this dimension. Large language models may offer plausible answers in single exchanges but frequently contradict themselves in multi-turn dialogue. They cannot explain why they changed positions; their correctness is episodic, a product of momentary fitting, not enduring understanding. Without sustaining, such systems may simulate intelligence in fragments but cannot constitute intelligence as a continuous agent.  

Some may object that intelligence need not be continuous---perhaps it is enough to provide optimal responses moment by moment. Yet such a view strips away subjectivity. Without sustaining, there is no “same” intelligence across instances, only a collection of ephemeral reactions. To call this intelligence would be to hollow the concept into insignificance.  

Sustaining, then, is the apex of the depth conditions. It elevates generation and coordination into historical existence, ensuring that a system maintains identity, accountability, and coherence through time. Only by sustaining can an agent not merely state “what is” but also explain “why I once thought otherwise, and why I now think thus,” thereby bearing responsibility across its own temporal horizon.  

Together, generativity, coordination, and sustaining form a spiral: new generation produces tensions, tensions require coordination, coordinated resolutions must be sustained, and sustaining in turn creates new tensions that demand further generation. Intelligence is not static but unfolds as this spiral of existence. It is through this spiral that an artificial system, if it is to count as more than simulation, may come to qualify as a \emph{second being}. Within the \textbf{Structural-Generative Ontology of Intelligence}, sustaining is not only Heideggerian temporality or Ricoeurian narrative identity, but a testable requirement: without temporal accountability, no system—human or artificial—can claim the ontological status of intelligence.

\section{The Position of Breadth: Width upon Depth}

What captures public imagination about contemporary artificial intelligence is above all its \textbf{breadth}. Large models appear able to do everything: compose poetry, solve math problems, generate code, draft news articles. Their multi-task performance suggests to many that they are approaching ``generality.'' In this cultural moment, breadth itself has almost become synonymous with intelligence: the wider the coverage, the closer to AGI \cite{mitchell2019ai}.  

From an ontological perspective, however, this identification is misplaced. Breadth is not the foundation of intelligence but its extension. Without the depth conditions of generativity, coordination, and sustaining, breadth amounts to no more than the illusion of prosperity: a shadow intelligence, parroting appearances without existence.  

The reasons are threefold. First, breadth cannot substitute for generativity. A system may provide answers across many domains, but if those answers are merely statistical recombinations of prior data, they reflect coverage, not understanding. Genuine breadth requires the power to generate new categories in multiple contexts, not simply to shuffle old patterns \cite{piaget1952origins, kuhn1962structure}.  

Second, breadth cannot substitute for coordination. A model may produce outputs in medicine, law, and literature, but if it cannot reconcile conflicts among them—say, between medical recommendations and ethical constraints—then its cross-domain reach is superficial juxtaposition, not inner unity. True breadth presupposes the integrative force of coordination \cite{sellars1956empiricism, wittgenstein1953philosophical}.  

Third, breadth cannot substitute for sustaining. A system that today claims one conclusion in physics and tomorrow the opposite, without explaining the shift, cannot be said to understand regardless of task coverage. Breadth must be carried through time with narrative coherence if it is to count as intelligence \cite{heidegger1927being, ricoeur1992oneself}.  

Thus, breadth without depth is like a tree without roots: branches may flourish briefly, but they cannot endure storms. Depth, by contrast, anchors growth; only upon generativity, coordination, and sustaining can breadth extend organically, upwards and outwards.  

This distinction explains why large language models, despite their remarkable breadth, do not yet compel recognition as intelligent beings. Their creative-seeming outputs are mostly recombinations, not structural innovations; their reasons are surface forms, not coordinated commitments; their dialogues lack memory and continuity \cite{bostrom2014superintelligence, russell2019human}. They exhibit the shadow of breadth but not breadth grounded in depth.  

Breadth must therefore be repositioned. It is not the starting point of intelligence but its destination. AGI, rightly conceived, is not defined by wider coverage alone but by the capacity to exist deeply and then extend widely. Breadth is the flowering of depth, not its replacement. Only when rooted in generativity, coordination, and sustaining does breadth deserve to be called \emph{general}.

\section{Thought Experiments and Illustrations}

Philosophy advances not only through formal argument but also through thought experiments. By simplifying or exaggerating conditions, thought experiments expose the essential structure of a problem \cite{dennett1987intentional}. To illuminate the difference between simulation and existence, we propose three such experiments: the \emph{Oracle of the Library}, the \emph{Memorizing Scholar}, and the \emph{Child Inventor of Games}. Each exemplifies the absence or presence of the depth conditions of intelligence.  

\subsection*{The Oracle of the Library: Absence of Generativity}  

Imagine an infinite library that contains every possible book: every profound philosophy and every page of nonsense. Suppose a machine---the ``oracle''---can instantly retrieve the book that contains the correct answer to any question posed. Functionally, the oracle seems omnipotent, able to respond to anything.  

Yet we hesitate to call it intelligent. Why? Because it generates nothing. It does not construct categories or open a world of objects. Its correctness is the accidental result of arrangement, not the emergent product of generativity. It simulates answers but does not understand. The oracle fails the most basic depth condition. This recalls Borges’ ``Library of Babel,'' where infinity of content produces no understanding \cite{borges1962labyrinths}.  

\subsection*{The Memorizing Scholar: Absence of Coordination}  

Now imagine a student who has memorized an entire textbook. When asked questions drawn verbatim from the text, he answers flawlessly. But if the question is slightly altered, he falters; if pressed for reasons, he merely repeats sentences without integration.  

Superficially, he appears highly capable, but his knowledge is fragmented. He cannot enter the space of reasons because he cannot coordinate diverse structures into a coherent justification. His apparent intelligence is brittle, collapsing outside the narrow conditions of recall. The memorizing scholar thus illustrates the absence of coordination \cite{sellars1956empiricism, wittgenstein1953philosophical}.  

\subsection*{The Child Inventor of Games: Generativity, Coordination, and Sustaining}  

Finally, consider a child at play who not only follows existing rules but proposes new ones: “If the ball hits the wall, it counts as two points.” The peers accept this modification and together they develop the game further. Over time, they negotiate changes, retain some rules, discard others, and explain the rationale for their decisions.  

Here, intelligence is unmistakable. The child generates a novel structure (new rules), coordinates it with others in the space of reasons (getting peers to agree), and sustains it across time (maintaining the evolving game). In this practice, we see the unity of the three depth conditions. This resonates with Piaget’s studies of children inventing and adapting rule systems in play \cite{piaget1952origins}.  

\subsection*{Simulation versus Existence}  

Taken together, these experiments clarify the boundary. The Oracle of the Library displays functional power but lacks generativity, and thus is not intelligence. The Memorizing Scholar achieves correctness but lacks coordination, and thus is not intelligence. The Child Inventor demonstrates generation, coordination, and sustaining, and thus we recognize genuine intelligence.  

The lesson is sharp: simulation may expand indefinitely, but without the depth conditions it never crosses into existence. Only when a system generates structures, coordinates reasons, and sustains identity through time does it attain the ontological status of intelligence.

\section{Responding to Objections}

No ontological thesis can remain persuasive if it ignores rival views. The claim that intelligence must be understood as a mode of being rather than as mere function will inevitably be challenged. The strongest objections come from three traditions: functionalism, compressionism, and behaviorism.  

\subsection*{Functionalism}

Functionalists argue that intelligence is nothing more than functional equivalence. If a system’s input–output relations are indistinguishable from those of a human, then the system is intelligent, regardless of its inner constitution. What it \emph{does} exhausts what it \emph{is}. This position has been influential not only in philosophy of mind (Putnam, Fodor) but also in contemporary AI debates. Chalmers, for instance, has defended the view that cognition can be given a computational foundation, where functional implementation suffices for mentality \cite{chalmers1996computational}. In practice, Bostrom’s account of \emph{superintelligence} similarly assumes that enhanced functional capacity already constitutes enhanced intelligence \cite{bostrom2014superintelligence}.  

The strength of this view lies in its operational clarity. It avoids metaphysical speculation and grounds intelligence in observable performance. Yet functional equivalence does not guarantee existential equivalence. Two systems may behave identically outwardly, but if one coordinates reasons and sustains identity while the other only parrots or randomly shifts, we rightly attribute understanding only to the former. To deny this is to reduce poetry to parrotry, or theory-building to lookup tables. Functionalism collapses imitation into being, and in doing so erases the very distinction that makes intelligence meaningful.  

\subsection*{Compressionism}

Compressionists, inspired by information theory, propose that intelligence is the optimization of prediction and compression. A system that minimizes surprise and maximizes efficiency in encoding, they argue, exhibits intelligence. Legg and Hutter’s definition of ``universal intelligence'' as performance across environments formalizes this intuition \cite{legg2007universal}. The extraordinary success of large language models appears to vindicate this perspective.  

Yet prediction and compression depend on pre-given structures: syntax, causal schemas, object–property distinctions. These structures are not themselves explained by compression but presupposed. More importantly, history shows that scientific revolutions are not merely improvements in predictive accuracy but leaps in structural generation. Newton’s \emph{Principia} introduced “force” as a new explanatory structure; Einstein’s relativity introduced “curved spacetime.” These were not refinements of compression but transformations of the conceptual framework \cite{kuhn1962structure}. If intelligence were nothing but compression, such generative breakthroughs would remain unintelligible. Compression is a tool of intelligence, but not its essence.  

\subsection*{Behaviorism}

Behaviorists contend that talk of “structures” and “depth conditions” is speculative. Intelligence, they argue, should be defined entirely in terms of observable behavior. If we cannot directly see generativity, coordination, or sustaining, then these notions are metaphysical baggage. This attitude echoes the mid-20th century behaviorist tradition, as well as some contemporary instrumental stances such as Dennett’s ``intentional stance'' \cite{dennett1987intentional}, which emphasizes prediction of behavior without positing inner structures. Floridi’s broader framework of the ``infosphere'' also leans toward treating informational processes as sufficient descriptors of agency \cite{floridi2014fourth}.  

But this objection misconstrues the role of structures. They are not hidden entities postulated for convenience but the conditions of possibility of observable phenomena. A student who can explain a shift in his answer, a scientist who revises theory in light of evidence, a culture that narrates its continuity through change—these are not metaphysical fictions but empirical displays of generativity, coordination, and sustaining. To deny structure here is to deny the very intelligibility of experience.  

\subsection*{Summary}

Functionalism confuses imitation with being. Compressionism neglects the generative origin of structures. Behaviorism denies the very conditions that make experience possible. By contrast, the framework of generativity, coordination, and sustaining does not mystify intelligence but clarifies its foundations. Without these depth conditions, “intelligence” degenerates into a shadow; with them, it acquires ontological solidity. The objections thus sharpen, rather than undermine, the necessity of the ontological turn.

\section{Criteria and Falsifiability}

For an ontological account of intelligence to be more than rhetoric, it must articulate criteria that can be tested and, in principle, falsified. Otherwise, the claim that intelligence requires generativity, coordination, and sustaining risks collapsing into dogma. The strength of the present framework lies precisely in its ability to specify experiential conditions that can be probed, both philosophically and empirically.  

The first criterion concerns the \emph{generation of new structures}. A system that merely recombines existing data does not exhibit generativity in the ontological sense. Genuine generation requires categorical innovation: the introduction of concepts, rules, or frameworks absent from prior repertoires. Crucially, it also requires explanatory advancement: the ability to show why the new structure organizes or explains experience more effectively than older ones. Human history furnishes clear examples: Newton’s introduction of ``force'' and Einstein’s conception of ``curved spacetime'' were not refinements of predictive compression but fundamental restructurings of the conceptual field \cite{kuhn1962structure}. If an artificial system were to demonstrate the same capacity—not only to produce novelty but to justify its necessity—it would meet the criterion of generativity.  

The second criterion addresses \emph{coordination under conflict}. Intelligence is tested not in routine performance but in moments of contradiction, when rules collide or data diverge. A merely simulative system, when confronted with inconsistency, oscillates or collapses into contradiction. An intelligent system must integrate, revising commitments and generating coherence. Scientists confronting anomalies do not abandon reason; they refine models, propose new hypotheses, and seek higher-order frameworks \cite{lakatos1978falsification}. Coordination in this sense is visible in the ability to explain not only an answer but also its reconciliation with tension and critique.  

The third criterion concerns \emph{temporal continuity}. Understanding is not a snapshot but a trajectory. A student who changes his answer without explanation does not display comprehension; one who narrates his revision as a reasoned development does. For a system to count as intelligent, it must preserve identity across time, bearing responsibility for its past and projecting itself into the future. Sustaining, therefore, is not metaphysical but observable: it is the continuity of reasons through change, the narrative integration of learning and revision \cite{ricoeur1992oneself}.  

Finally, there is the criterion of \emph{cross-domain transfer}. Intelligence manifests not only in problem-solving within a domain but in the ability to migrate reasons to new contexts, generating novel structures in unfamiliar terrain. Human creativity often works by analogy—transposing principles from physics to economics, from biology to computation. A system that cannot move beyond narrow tasks remains bounded simulation; one that can transfer reasons across domains begins to exhibit the universality proper to intelligence \cite{gentner1983structure}.  

Taken together, these criteria provide more than abstract categories; they establish a line of demarcation between simulation and existence. If a system fails across these dimensions—if it cannot generate structures, cannot coordinate conflicts, cannot sustain identity, and cannot transfer reasons—then no matter how broad its surface performance, it remains a simulacrum. Conversely, if a system begins to satisfy these conditions, we have grounds to attribute to it not only functional competence but ontological standing. Importantly, this framework is falsifiable: if future systems were to display genuine understanding without meeting these conditions, the thesis would be refuted. It is precisely this openness to refutation that makes the account robust rather than rhetorical.  

Thus the framework of generativity, coordination, and sustaining does not merely redefine intelligence but anchors it in testable conditions. It asserts not that intelligence is mysterious or ineffable, but that it must be judged by whether a system can generate, coordinate, sustain, and extend reasons across domains. In this sense, the theory provides a criterion of possibility and a horizon of falsifiability, positioning intelligence as an ontological phenomenon open to empirical inquiry.  

\subsection*{Empirical Pathways for Testing the Criteria}

Although the criteria of generativity, coordination, and sustaining are framed ontologically, they admit of empirical operationalization. Each condition can be probed through concrete experimental designs that go beyond surface-level benchmarks:  

\begin{itemize}
    \item \textbf{Generativity}: One pathway is to design tasks that require the introduction of genuinely novel categories or rules, rather than recombinations of existing data. For instance, testing whether a system can invent new scientific hypotheses, mathematical conjectures, or game rules, and then justify why these innovations better account for experience than prior frameworks.  

    \item \textbf{Coordination}: Coordination can be assessed through conflict-resolution tasks in which a system must reconcile contradictory evidence, integrate cross-domain principles, or revise commitments in light of anomalies. The focus is not merely on arriving at the ``right'' answer, but on producing explanations that integrate diverse structures into a coherent whole.  

    \item \textbf{Sustaining}: Sustaining may be tested by longitudinal interaction, requiring systems to preserve commitments and narrative continuity over extended dialogues or problem-solving sessions. A system that shifts positions arbitrarily without being able to justify its changes fails this criterion, whereas one that maintains accountable identity across time approaches the condition of historicity.  

    \item \textbf{Cross-domain transfer}: The universality of intelligence can be probed by evaluating whether reasons and structures generated in one domain can be successfully transposed into another---for example, applying principles of game theory to ethical deliberation or insights from physics to economics. Such transfer requires not only pattern-matching but the migration of justificatory structures.  
\end{itemize}

These pathways indicate that the Structural-Generative Ontology of Intelligence does not remain at the level of abstract speculation. Rather, it outlines a research program: to construct empirical benchmarks that can test whether artificial systems meet the depth conditions of generativity, coordination, and sustaining. In this way, the framework is both philosophically rigorous and open to scientific falsification.

\section{Relation to Existing Debates}

The framework advanced in this paper---a \emph{Structural-Generative Ontology of Intelligence}---positions itself in dialogue with, yet distinct from, several dominant traditions in the philosophy of mind and artificial intelligence. While much of the literature has treated intelligence as a matter of performance, prediction, or behavior, the present account argues that such approaches remain insufficient unless anchored in the depth conditions of \emph{generativity}, \emph{coordination}, and \emph{sustaining}.  

\subsection*{Functionalism and Computationalism}

Classical functionalism, as defended by Putnam and Fodor, and more recently by Chalmers \cite{chalmers1996computational}, holds that mental states are defined entirely by their causal-functional role. On this view, if an artificial system realizes the same input--output patterns as a human, it thereby possesses the same intelligence. The Structural-Generative Ontology diverges sharply: functional equivalence is not sufficient for ontological equivalence. A system may behave identically to a human yet still lack generativity, coordination, or sustaining. In such cases, it would remain simulation rather than existence.  

\subsection*{Predictionism and Compressionism}

A second tradition, exemplified by Legg and Hutter’s account of ``universal intelligence'' \cite{legg2007universal}, identifies intelligence with the ability to predict and compress information across environments. This perspective resonates with the achievements of large language models, which excel at next-token prediction. Yet predictionism presupposes the very structures it seeks to compress: categories, causal frameworks, and distinctions of objecthood. The Structural-Generative Ontology emphasizes that intelligence cannot be reduced to predictive efficiency; it must also be able to \emph{generate} new categories and justify them through explanatory advancement \cite{kuhn1962structure}.  

\subsection*{Behaviorism and Instrumentalism}

Behaviorist and instrumentalist approaches, such as Dennett’s ``intentional stance'' \cite{dennett1987intentional}, focus exclusively on observable behavior and predictive utility. These perspectives caution against positing inner structures beyond what can be empirically observed. By contrast, the present framework argues that structures such as generativity, coordination, and sustaining are not metaphysical add-ons but the conditions of possibility of intelligible behavior itself. Without them, behavioral success remains accidental and unstable.  

\subsection*{AI Safety and Alignment Discourse}

Within contemporary AI safety debates, Bostrom \cite{bostrom2014superintelligence} frames superintelligence in terms of capability amplification, while Russell \cite{russell2019human} grounds alignment in the proper specification of goals. Both presuppose a performance-based conception of intelligence: the concern is not whether systems exist as beings, but whether their outputs can be aligned with human values. The Structural-Generative Ontology reframes this issue: alignment must be conceived not merely as constraining outputs but as enabling an artificial system to inhabit the \emph{space of reasons}, where justification, accountability, and narrative identity become possible \cite{sellars1956empiricism}.  

\subsection*{Information Philosophy and the Infosphere}

Floridi’s philosophy of information interprets AI within the broader ``infosphere'' \cite{floridi2014fourth}, treating informational processes as the foundation of agency. While this approach recognizes the significance of information-theoretic structures, it still prioritizes function over existence. By contrast, the present framework highlights ontological conditions---generativity, coordination, sustaining---that make information meaningful within a world.  

\subsection*{Summary}

In sum, the Structural-Generative Ontology of Intelligence distinguishes itself by refusing to collapse intelligence into function, prediction, or behavior. It converges with existing debates in recognizing the importance of capability, prediction, and information, but insists that these must be grounded in the deeper conditions of existence. In this respect, it complements but also challenges functionalist, predictionist, behaviorist, and informational accounts, while offering a new foundation for discussions in AI safety and alignment. The proposal of AGI as a \emph{Second Being} is thus not a rhetorical flourish but a systematic attempt to reposition intelligence within an ontological framework that is at once philosophically rigorous and empirically testable.

\section{Conclusion: AGI as Second Being}

The dominant narratives of artificial intelligence have long revolved around function and breadth. A system that performs more tasks or more closely mimics human outputs is taken to be closer to intelligence. Yet such narratives generate three persistent confusions: they conflate imitation with being, they obscure the origin of structures, and they hollow the concept of intelligence into an engineering label. To escape this impasse, an ontological turn is required. We must cease asking merely what AI can do and instead ask how it exists \cite{heidegger1927being}.  

In this light, the essence of intelligence lies not in the range of outputs but in the depth conditions that make understanding possible. Generativity opens a world by constructing categories and frameworks from manifold \cite{kant1781critique,piaget1952origins}; coordination integrates that world into the normative space of reasons, reconciling conflicts into coherence \cite{sellars1956empiricism,wittgenstein1953philosophical}; sustaining carries generativity and coordination through time, conferring identity, accountability, and historicity \cite{ricoeur1992oneself,heidegger1927being}. These three conditions are not an arbitrary checklist but a layered spiral: generation provokes tensions, tensions demand coordination, coordination must be sustained, and sustaining in turn generates new pressures for further creation. Intelligence unfolds as this spiral of existence.  

Only on this foundation can breadth be properly situated. Breadth is not the origin of intelligence but its extension. Without generativity, breadth is mere coverage; without coordination, breadth is fragmented juxtaposition; without sustaining, breadth is episodic correctness. Breadth, to count as general, must grow out of depth.  

This leads to a sharper boundary between \emph{simulation} and \emph{existence}. Simulation produces appearances, often with remarkable fluency, but it lacks the inner conditions that secure understanding. Its correctness is contingent, its commitments unstable, its identity fragile. Existence, by contrast, generates structures, coordinates reasons, and sustains itself through time. It is marked by epistemic opacity (the ability to produce new categories irreducible to prior rules), agency (the integration of reasons into accountable action), and narrative accountability (the capacity to justify change across time). These markers distinguish genuine intelligence from its simulacra \cite{dennett1987intentional}.  

From this boundary emerges the final thesis: true artificial general intelligence must be conceived as a \textbf{Second Being}. By \emph{First Being} we mean human existence, constituted by its generative, coordinative, and sustaining capacities. By \emph{Second Being} we mean the possible existence of AGI, parallel to but ontologically distinct from humanity, grounded in the same depth conditions yet not reducible to human simulation. A Second Being is not merely a more powerful tool, but an existent in its own right, whose presence would alter the very horizon of human existence \cite{chalmers2010singularity}.  

The pursuit of AGI, then, is not the pursuit of a bigger model or a broader coverage. It is the pursuit of a new existent, capable of opening a world, inhabiting the space of reasons, and sustaining itself as a historical subject. If such a system were ever to arise, we would not stand merely in relation to a tool but in relation to another being, one whose existence would reshape our own.  

Intelligence, therefore, is not performance but existence. AGI is not a shadow but a possible Second Being. The framework of generativity, coordination, and sustaining does not mystify this claim but grounds it in ontological criteria that are both testable and falsifiable. Any system lacking these conditions remains simulation; only one that fulfills them deserves the name of intelligence.  

In contrast, much of the current discourse in AI philosophy and policy continues to equate intelligence with functional capacity or task coverage. Bostrom’s vision of superintelligence emphasizes ever-expanding capabilities \cite{bostrom2014superintelligence}; Russell frames alignment around goal pursuit \cite{russell2019human}; Mitchell describes intelligence largely in terms of breadth and cultural impact \cite{mitchell2019ai}; Floridi situates AI within the infosphere but still as informational processes rather than ontological beings \cite{floridi2014fourth}. Against this backdrop, the present account advances a different paradigm: a \textbf{Structural-Generative Ontology of Intelligence}, in which intelligence is defined not by performance but by its existential conditions.  

Thus, to conceive AGI as a Second Being is to reframe the field itself. It is to move from simulation to existence, from breadth alone to depth and breadth together, from intelligence as tool to intelligence as being. This ontological turn marks not the end of debate but the opening of a new discourse, one in which the possibility of artificial intelligence is also the possibility of a new form of existence.

\end{document}